\documentclass{article}

\usepackage{microtype}
\usepackage{graphicx}
\usepackage{subfigure}
\usepackage{booktabs} 
\usepackage{hyperref}

\usepackage[accepted]{icml2023_dp4ml}

\usepackage{amsmath}
\usepackage{amssymb}
\usepackage{mathtools}
\usepackage{amsthm}

\usepackage[capitalize,noabbrev]{cleveref}

\theoremstyle{plain}

\theoremstyle{definition}

\theoremstyle{remark}

\usepackage[textsize=tiny]{todonotes}
\newtheorem{thm}{Theorem}

\newtheorem*{prop*}{Proposition}

\theoremstyle{definition}

\DeclareMathOperator{\Tr}{Tr}

\DeclareMathOperator{\diag}{diag}

\newcommand{\argmax}[1]{\underset{#1}{\operatorname{arg}\,\operatorname{max}}\;}
\newcommand{\argmin}[1]{\underset{#1}{\operatorname{arg}\,\operatorname{min}}\;}

\renewcommand{\a}{{\bf a}}

\renewcommand{\c}{{\bf c}}

\newcommand{\h}{{\bf h}}

\newcommand{\w}{{\bf w}}
\newcommand{\x}{{\bf x}}
\newcommand{\y}{{\bf y}}
\newcommand{\z}{{\bf z}}

\newcommand{\I}{{\bf I}}

\newcommand{\M}{{\bf M}}

\newcommand{\R}{\mathbb{R}}

\newcommand{\W}{{\bf W}}
\newcommand{\X}{{\bf X}}
\newcommand{\Y}{{\bf Y}}
\newcommand{\Z}{{\bf Z}}

\newcommand{\Real}{{\mathbb R}}

\icmltitlerunning{Duality Principle and Biologically Plausible Learning}

\begin{document}

\twocolumn[
\icmltitle{Duality Principle and Biologically Plausible Learning: \\ Connecting the Representer Theorem and Hebbian Learning}



\icmlsetsymbol{equal}{*}

\begin{icmlauthorlist}
\icmlauthor{Yanis Bahroun}{yyy}
\icmlauthor{Dmitri B. Chklovskii}{yyy,comp}
\icmlauthor{Anirvan M. Sengupta}{yyy,sch}
\end{icmlauthorlist}

\icmlaffiliation{yyy}{Flatiron Institute, Simons Foundation, NY, USA}
\icmlaffiliation{comp}{Department of Physics and Astronomy, Rutgers University, NJ, USA}
\icmlaffiliation{sch}{Neuroscience Institute, NYU Langone Medical Center, NY, USA}

\icmlcorrespondingauthor{Yanis Bahroun}{ybahroun@flatironinstitute.org}

\icmlkeywords{Machine Learning, ICML}

\vskip 0.3in
]



\printAffiliationsAndNotice{\icmlEqualContribution} 

\begin{abstract}
A normative approach called Similarity Matching was recently introduced for deriving and understanding the algorithmic basis of neural computation focused on unsupervised problems. 
It involves deriving algorithms from computational objectives and evaluating their compatibility with anatomical and physiological observations. 
In particular, it introduces neural architectures by considering dual alternatives instead of primal formulations of popular models such as PCA. 
However, its connection to the Representer theorem remains unexplored. 
In this work, we propose to use teachings from this approach to explore supervised learning algorithms and clarify the notion of Hebbian learning. 
We examine regularized supervised learning and elucidate the emergence of neural architecture and additive versus multiplicative update rules.
In this work, we focus not on developing new algorithms but on showing that the Representer theorem offers the perfect lens to study biologically plausible learning algorithms. 
We argue that many past and current advancements in the field rely on some form of dual formulation to introduce biological plausibility. 
In short, as long as a dual formulation exists, it is possible to derive biologically plausible algorithms. 
Our work sheds light on the pivotal role of the Representer theorem in advancing our comprehension of neural computation.
\end{abstract}

\section{Introduction}
\label{introduction}

Machine learning (ML) and neuroscience have influenced each other in many ways \cite{bengio2015towards,guerguiev2017towards}. 
In particular, the workings of the human brain have inspired the development of ML algorithms \cite{fukushima1980neocognitron}, in particular artificial neural networks (NN). 
At the same time, neuroscience has benefited from the insights and models developed in ML \cite{schrimpf2018brain}.

One of the main drivers of biological plasticity is the Hebbian principle \cite{hebb1949organization}. 
It is based on the idea that the strength of a connection between neurons increases when they are activated together. 
Hebbian learning models many different aspects of learning in the brain, and numerical implementations of it are known to learn reoccurring activity patterns \cite{oja1982simplified,rumelhart1985feature,sanger1989optimal,olshausen1996emergence}. 
An algorithm using such rules can process and adapt to one input at a time with memory requirements independent of the number of samples. 
Indeed, inspired by the Hebbian principle, the contributions above enabled substantial developments in ML and the understanding of neural mechanisms. 
Regrettably, they are often based on ad hoc learning rules, and there are few rigorous formulations.

The Representer theorem \cite{wahba1990spline,scholkopf2001generalized,argyriou2009there} is a crucial component for connecting standard and kernel formulations of ML algorithms. 
The theorem states that the optimal solution to a particular class of problems can be expressed as a linear combination of the training samples in a feature space. 
The Representer theorem has been used to develop efficient algorithms for solving many ML problems, including regression, classification, and clustering. 
Recently, kernel methods have been omnipresent for studying deep NNs, either from NTK \cite{jacot2018neural,arora2019exact} or using techniques from statistical mechanics \cite{bordelon2020spectrum,canatar2021spectral}. 
However, although the kernel approach to artificial NNs has provided insights into the inner workings of artificial NNs, it is unclear how they can benefit our understanding of biological ones.

This paper highlights that primal-dual formulations of regularized models provide a perfect framework for understanding the relationships between Hebbian learning and the Representer theorem. 
More precisely, by connecting regularized models with neural architectures in the primal space, Hebbian learning rules in the dual space, and the primal-dual with neural dynamics, the theory of duality provides a powerful tool for understanding the mechanisms of learning in ML and neuroscience. 
The link between Hebbian learning and the Representer theorem is an example of the deep insights that can be gained by exploring the relationships between these two fields. 
In practice, this can also lead to developing local and biologically plausible alternatives to algorithms, such as backpropagation (BP) or neuromorphic implementation of powerful deep NNs. 
This work aims at extending the existing work from Similarity Matching (SM) to supervised learning algorithms. The work of \citet{pehlevan2018similarity} first focused on connecting Hebbian learning rules and similarity-based objective functions.

\textbf{Outline.}
In Section~\ref{sec:background}, we describe the background of Hebbian learning, synaptic plasticity, and a recently developed framework called Similarity Matching (SM). SM borrows from the duality theory to derive biologically plausible algorithms for unsupervised learning. 
In Section~\ref{sec:Duality}, we recall the theory of duality for models with regularized objectives with convex risks, the key aspect being the existence of the Representer theorem. 
In Section~\ref{sec:NNandPlasticity}, we explore the connection between neural dynamics, Hebbian rules, and the dual formulation of commonly used models, e.g., Ridge regression, SVM, Logistic regression, and Entropy-regularized problems. 
In Section~\ref{sec:Connection}, we further explain the consequences and advantages of this method for studying learning algorithms. 
Finally, in Section~\ref{sec:Discussion}, we discuss the implication and how it elucidates various methods used in recent years to develop biologically plausible and local alternatives to BP.

\section{Background}\label{sec:background}

In recent years, there has been a growing interest in building local and biologically plausible alternatives to the popular BP algorithm \cite{rumelhart1986learning} for several reasons.

While highly effective for training DNNs, BP relies on the computation of gradients across all network layers, which is not biologically plausible. 
Moreover, BP requires extensive computations and information exchange throughout the network, which can be computationally expensive and energy-consuming. 
Finally, BP typically requires access to large batches of data for practical training, making it less suitable for online learning scenarios.

Biological systems must instead rely on local computations, are energy-efficient, and are trained online. 
As a result, developing algorithms that align with the constraints and mechanisms observed in biological systems is important for understanding neural processes and potentially uncovering novel learning mechanisms. 
It can also lead to more efficient and scalable learning processes and enable continual learning.
Such developments could also be crucial for the further development of neuromorphic computing.

\subsection{Hebbian learning and neural dynamics}

The notion of local and online learning rules is essential in neuroscience because it reflects the idea that locally available quantities drive changes in neural connections and activity.

\paragraph{Hebbian rule.}
The Hebbian principle \cite{hebb1949organization} is a famous local learning rule stating that if a pre-synaptic neuron $x_i$, and a post-synaptic neuron, $z_j$, are active at the same time, the strength of the connection between them, denoted by $w_{ij}$ should increase as 
\begin{align}\label{eq:basic_hebbian_rule}
\Delta w_{ij} \propto \eta x_i z_j ~~,
\end{align}
where $\Delta w_{ij}$ is the change in the weight, and $\eta$ a learning rate. 
The Hebbian principle is important because it provides a mechanism for neurons to learn associations between stimuli and responses, a fundamental aspect of many types of learning and memory. 
We do not discuss the limitation of this principle here, and we direct the reader toward the literature on BCM \cite{bienenstock1982theory} or STDP \cite{sjostrom2001rate} if interested.

\paragraph{Neural dynamics.}
Without being a dynamical systems approach to neuroscience, we assume that neural activity is a function of its input and not a clipped version of it. The solution is obtained by running the following dynamical systems until convergence
\begin{align}\label{eq:basic_neural_dynamics}
    \frac{dz_t(\gamma)}{d\gamma}  = \Gamma \left(\w,\x_t,y_t,z_t(\gamma)\right)~~,
\end{align}
where $\Gamma(\cdot)$ is a possibly nonlinear function, and $\gamma$ the steps in the dynamics.

\subsection{Similarity matching framework}\label{subsec:SM}

Interestingly, in the SM framework, unlike in other models for biologically plausible models, the authors start from the dual problem, which they solve in the primal by introducing auxiliary variables. 
\paragraph{Dual problem.} 
In one of the first papers on SM \cite{pehlevan2015hebbian}, the authors borrowed the objective from Classical Multidimensional scaling \cite{cox2000multidimensional} (CMDS), using the equivalence that exists between the solution of the CMDS and kernel PCA \cite{scholkopf1998nonlinear} with linear kernel. 
The objective function that the authors propose to solve is the following. Given an input data matrix $\X = [\x_1,\ldots,\x_T] \in \Real^{n\times T}$, find an output matrix $\hat{\Z} \in \Real^{m\times T}$, with $m \leq n$, such that 
\begin{align}\label{eq:CMDS}
    \hat{\Z} = \argmin{\Z}  - 2 \Tr( \X^\top \X \Z^\top \Z) + \| \Z^\top \Z \|_F^2 ~~.
\end{align}
The $\hat{\Z}$ variables, in Eq.~\eqref{eq:CMDS}, represent the dual representation of the weight parameters, $\W$ defined in PCA \cite{williams2001connection} as 
\begin{align}
    \hat{\W} = \argmax{\W \in \Real^{m\times n}} \frac{1}{T} \Tr\left( \W \X \X^\top \W^\top \right)
\end{align}
under the orthonormality constraint $\W\W^\top = \I_m$, with $m\leq n$. 
This result derives from the duality relationship that exists between the eigenvalue decomposition of the covariance matrix $\X\X^\top$ and that of the Gram matrix $\X^\top \X$. 
An issue with the dual formulation is that there is no obvious formulation in terms of synaptic plasticity.

\paragraph{Primal-dual formulation.}
The authors of \cite{pehlevan2015normative,pehlevan2018similarity} proposed to solve the dual problem, Eq.~\eqref{eq:CMDS}, by reformulating it as the following primal-dual min-max optimization problem:
\begin{align}\label{eq:minmax_SM}
    \min_{\Z}\max_{\M}\min_{\W} -2 \Tr(\Z^\top\W\X) + \Tr(\Z^\top\M\Z) \nonumber \\
    + \Tr(\W^\top\W) - \tfrac{1}{2}\Tr(\M^\top\M) ~~.
\end{align}
The reformulated problem Eq.~\eqref{eq:minmax_SM} did not introduce one but two auxiliary (primal) variables $\W$ and $\M$, which have different neural interpretations. 

\paragraph{Neural implementation.}
The resulting model exhibits a different architecture than primal-based models \cite{oja1992principal}, as it incorporates lateral inhibitory connections encoded by the weight matrix $\M$. 
The neural dynamics of the resulting algorithm and update rules are obtained by gradient optimization applied to the objective function defined in Eq.\eqref{eq:minmax_SM} as follows
\begin{align}
    \text{Neural dynamics:}~~ &\frac{d\Z(\gamma)}{d\gamma}  = \W\X - \M\Z(\gamma)~, \label{eq:Ndyn_SM} \\
    \text{Update rules:}~~ & \Delta \W  = \Z\X^\top ~,~ \Delta \M  = - \Z\Z^\top ~. \label{eq:update_SM}
\end{align}
The neural dynamics Eq.~\eqref{eq:Ndyn_SM}, and the update rules  Eq.~\eqref{eq:update_SM} respectively satisfy the basics definition for biological plausibility defined in Eq.~\eqref{eq:basic_neural_dynamics} and Eq.~\eqref{eq:basic_hebbian_rule}.

In a series of recent works \citep{bahroun2019similarity,lipshutz2020biologically,lipshutz2021biologically,bahroun2021normative,lipshutz2022single}, the authors extended the SM framework to include objectives for more complex learning tasks. 
Examples include Canonical Correlation Analysis (CCA) \cite{hotelling1936relations}, Slow Feature Analysis \cite{wiskott2002slow}, and Independent Component Analysis \cite{hyvarinen2000independent}, which are all generalized eigenvalue (GEV) problems. 
Given that GEVs admit dual formulations, the algorithms derived from these objectives naturally map onto NNs with local forms of plasticity. 
Reviews of the SM framework can be found in \cite{pehlevan2019neuroscience,lipshutz2023normative}.

\section{Duality principle for regularized models}\label{sec:Duality}

The SM framework mentioned above leverages the primal-dual equivalence of problems to develop biologically plausible learning rules for unsupervised learning problems. 
In this work, the question is about the implication of applying such an idea to supervised learning problems. 
In this section, we recall the duality framework in a supervised setting.

\subsection{Regularized systems with convex risks}

Although the set of models considered can be more general, for practical relevance, we consider linear (possibly point-wise nonlinear, e.g., ReLU) prediction problems such as linear regression and classification. 
In the following, we restate the more general case and then decompose the problem in determining the dual for the loss and the regularizer, respectively. 
For a detailed presentation of the theory of duality and convex optimization, we invite the reader to refer to the work of \citet{scholkopf2001generalized} and \citet{zhang2002dual}.

We first set the notations. The data corresponds to samples $\x_t \in \mathcal{X} = \mathbb{R}^n$ and associated labels $\y_t \in \mathcal{Y}$, $\forall t \in \{1,\ldots,T\}$.  
The models we consider in the following are those of regularized systems with convex loss. 
Supervised learning problems are often phrased as minimizing an objective function with respect to a set of parameters $\w \in \mathcal{X}$ as 
\begin{align}\label{eq:gen_lin}
\hat{\w} = \argmin{\w \in \mathbb{R}^n} \left[ \frac{1}{T} \sum_{t=1}^T \ell \left(\y_t , f(\w^\top \x_t)\right) + \lambda g(\w) \right] ~.
\end{align}
The general assumptions are the following, $\ell(a, b)$ is a convex function of $b$, $g(\cdot)$ is a convex function of $\w$, and $\lambda>0$ is a regularization parameter. 
Given this setup, we recall below the Representer theorem.

\begin{thm}\label{thm:Representer_thm}
\textbf{Representer Theorem} \cite{scholkopf1998nonlinear} 
Suppose we are given a non-empty set $\mathcal{X}$, a positive real-valued kernel $k$ on $\mathcal{X}\times\mathcal{X} $, training sample $(\x_1, y_1), \ldots , (\x_T, y_T) \in \mathcal{X} \times \mathbb{R}$, a strictly monotonically increasing real-valued function $g$ on $[0,+\infty[$, an arbitrary cost function $\ell : \Real \times \Real \rightarrow \mathbb{R} \cup \{\infty\}$, and a class of functions 
\begin{align}
    \mathcal{F} = \left\{ f \in \Real^{\mathcal{X}}  | f(\cdot) = \sum_{i=1}^{\infty} \beta_i k(\cdot,\h_i), \beta_i \in \Real, \right. \nonumber \\ 
    \left. \h_i \in \mathcal{X}, \| f \| <\infty \right\}~.
\end{align} 
Here, $\|\cdot\|$ is the norm in the RKHS $H_k$ associated with $k$. 
Then any $f \in \mathcal{F}$ minimizing the regularized risk functional
\begin{align}
\frac{1}{T}\sum_{t=1}^T \ell (y_t, f(\x_t)) + \lambda g (\|f\|) ~,
\end{align}
admits a representation of the form 
\begin{align}
f(\cdot) = \sum_{t=1}^T z_t k(\cdot,\x_t)~~. 
\end{align}
\end{thm}

Although there is potentially a limitless amount of such objectives, it appears that the literature often revolves around the following few. 
The square loss, $\ell(\y,f(\w^\top \x)) = \frac{1}{2} \|\y - f(\w^\top \x)\|^2$, for regularized linear regression. 
The SVM loss, $\ell(\y,f(\w^\top \x)) = \ell (\y f(\w^\top \x)) = 1_{\y f(\x)<0}$, and cross-entropy as in logistic regression $\ell(\y,f(\w^\top \x)) = - \y \log(f(\w^\top \x)) - (1-\y) \log(1-f(\w^\top \x))$ for classification. 
For regression tasks consider $\mathcal{Y}= \mathbb{R}$ or $\mathbb{R}^n$, and instead  $\mathcal{Y}=\{-1,1\}$ for classification tasks.

\subsection{Dual formulation of the loss term $\ell(\cdot)$}

We first address the dual of the loss term, which is not only a function of the parameters, $\w$, and the samples, $\x$, but also of the labels $\y$. 
Let $c(y,\cdot)$ be the dual transform of $\ell (y,\cdot)$ defined as 
\begin{align}
c(y,v) = \sup_{u} (uv - \ell (y,u)),
\end{align}
implies that for all $t \in \{1,\ldots,T\}$
\begin{align}
\ell ( y_t, \w^\top \x_t  ) = \sup_{z_t} \left( z_t \w^\top \x_t  - c(y_t,-z_t) \right). 
\end{align}
Using this definition, we can now introduce an auxiliary, dual variable $\z$, with component $z_t$, such that the following objective
\begin{align}
R(\w, \z) = \frac{1}{T} \sum_{t=1}^T \left (-c(y_t,-z_t) - z_t \w^\top  \x_t \right) + \lambda g(\w), 
\end{align}
and the optimization problem, Eq.~\eqref{eq:gen_lin}, are equivalent the optimal solution $\hat{\w}$ can be equivalently rewritten as 
\begin{align}\label{eq:minmax}
\hat{\w} = \arg \inf_{\w} \sup_{\z} R(\w, \z).
\end{align}
Switching the order of $\inf_\w$ and $\sup_{\z}$ in the above minimax convex-concave programming problem is valid. 
Given $\hat{\w}$ a solution of the problem Eq.~\eqref{eq:minmax}, there is a solution $\z$ to the following problem
\begin{align}\label{eq:maxmin}
\hat{\z} = \arg \sup_{\z} \inf_{\w}  R(\w, \z). 
\end{align}

\subsection{Dual of the regularizer $g(\cdot)$}

The goal now is to have a formulation of Eq~\eqref{eq:maxmin} only as a function of $\z$. 
To do so, we now consider $h(\cdot)$ the dual transform of $g(\cdot)$ the regularizer defined by
\begin{align}
h(v) = \sup_{u} (u^\top v - g(u))~, 
\end{align}
where $h(\cdot)$ is also a convex function. Using the dual formulation of the regularizer, the optimization problem Eq.~\eqref{eq:maxmin} can be rewritten as the following dual formulation 
\begin{align}
\hat{\z} = \arg \inf_{z} \left[\frac{1}{T}\sum_{t=1}^T c(y_t,-z_t) + \lambda h\left( \frac{1}{\lambda T} \sum_{t=1}^T z_t \x_t \right)  \right]~,
\end{align}
strictly in terms of $\z$.

\subsection{General implications}

As a result, we see that the optimal solution $\hat{\w}$ to the primal problem in Eq.~\eqref{eq:gen_lin} is given by
\begin{align}\label{eq:Rep_Hebb}
\hat{\w} = \nabla h \left( \frac{1}{\lambda T} \sum_{t=1}^T \hat{z_t} \x_t \right)~.
\end{align}
Looking at Eq.~\eqref{eq:Rep_Hebb}, it becomes clear that it greatly resembles the form of the Hebbian rule, Eq.~\eqref{eq:basic_hebbian_rule}, where the synaptic weights are a weighted sum of the samples.

Interestingly, we also have
\begin{align}\label{eq:Rep_NDyn}
\hat{z_t} = - \ell' (y_t,\hat{\w}^\top\x_t) ~~,
\end{align}
where $\ell'$ denotes a (sub)gradient of $\ell (a, b)$ with respect to $b$. 
The dual variables $z_t$, Eq.~\eqref{eq:Rep_NDyn}, now have clearly defined neural dynamics as the solution to an optimization problem. 
It is a function of the samples, weights, and labels as defined in Eq~\eqref{eq:basic_neural_dynamics}. 

To clarify this point, we study in the following section (Sec.~\ref{sec:NNandPlasticity}) some of the most widely used models for regression and classification. 
In Section~\ref{sec:Discussion}, we will discuss recent developments of biologically plausible learning alternatives to BP and show how heavily they rely on the Representer theorem often implicitly. 
We also describe in Section~\ref{sec:Discussion}, what canonical neural components are used in some of the more elaborate biologically plausible implementations.

\section{Neural dynamics and plasticity rule}\label{sec:NNandPlasticity}

Using the general formulation of the learning problem presented in the previous section, we recall the dual formulation of popular models, such as Ridge regression, SVM, and logistic regression. 
We present the implications of choosing a regularizer and a cost function to the form of the update rule, i.e., additive ($L_2$) or multiplicative (Entropy), and either product or difference of kernels of label and samples.

The purpose of this section is to lay the groundwork for identifying the critical component for ``biological interpretation".

\subsection{Ridge Regression}\label{subsec:RR}

We consider $\mathcal{X}\in \mathbb{R}^n$ and $\mathcal{Y}\in \mathbb{R}$, with $f: \mathbb{R} \mapsto \mathbb{R}$ the simple linear function $f(\w^\top \x) = \w^\top \x$. 
We use here the quadratic loss defined as follows
\begin{align}\label{eq:RR}
\min_{\w\in\mathbb{R}^n} \sum_{t=1}^T ( y_{t} - \w^\top \x_t )^2  + \frac{\lambda}{2}\w^\top \w~~.
\end{align}
The problem Eq.~\eqref{eq:RR}, being part of the framework Sec.~\ref{sec:Duality}, admits a dual formulation of the loss. 
We now introduce the variables $z_{t=1..T}\in \mathbb{R}$, which identify with a prediction error term $(y_{t} - \w^\top \x_t)$ as
\begin{align}
(y_t -\w^\top \x_t)^2 = \max_{z_t} z_t ( y_t - \w^\top \x_t) - \frac{1}{2} z_t^2 . 
\end{align}
As a result, we can rewrite the original objective as a min-max objective, Eq.~\eqref{eq:RR}
\begin{align}
\min_{\w}\max_{\z} R(\w,\z) = \min_{\w }\max_{z_{t=1..T}} \sum_{t=1}^T z_t ( y_{t} - \w^\top \x_t ) \nonumber
\\ -\tfrac{1}{2}\sum_{t=1}^T z^2_t  + \tfrac{1}{2} \w^\top \w .
\end{align}
We swap the order of optimization to obtain the optimal set of parameters, $\hat{\w}$, as a function of the dual variables, $z_t$ as 
\begin{align}
\hat{\w} =  \sum_{t=1}^T z_t \x_t~~ .    
\end{align}

\paragraph{Dual objective function.} 
By replacing $\hat{\w}$ we obtain the following dual objective function for linear regression as
\begin{align}
\max_{z_t} R(\hat{\w},z_t) = \max_{z_t}  \sum_{t=1}^T z_t y_{t} -  \frac{1}{2} \sum_{t'=1}^T  z_t z_{t'}  \x_{t'}^\top \x_t  \nonumber
\\ -\frac{1}{2}\sum_{t=1}^T z^2_t ~~.
\end{align}
With abuse of notations, $\diag(\z) = \Z$, we can rewrite the problem as 
\begin{align}\label{eq:MM_RR}
\max_{\Z} \Tr(\Y^\top \Z) - \frac{1}{2}\Tr (\X^\top \X \Z^\top \Z ) - \frac{1}{2}\Tr( \Z^\top \Z) ~~. 
\end{align}
%

\paragraph{Connection with Similarity Matching.}
The dual objective above shares a lot of similarities with the SM objective, Eq.\eqref{eq:CMDS}. 
Indeed, the synaptic weights, $\w$, arise from the kernel-to-kernel relationship between samples and neural output variable, $\Tr(\X^\top\X \Z^\top \Z)$. 
Interestingly, although here, the optimization is a maximization rather than a minimization, it is the same object that is considered. 
This naturally leads to the fact that the neural variables, $\Z$, in a supervised setting correspond to errors.


\paragraph{Architecture, plasticity, and neural dynamics.}

By construction of $z_t$ and $\w$, it is already clear that the plasticity of this connection is Hebbian and that we can derive neural dynamics for $z_t$ and plasticity rules as 
\begin{align}
&\text{Neural dynamics:} ~~~\tfrac{dz_t(\gamma)}{d\gamma} = (y_t - \w^\top \x_t) - z_t(\gamma), \\
&\text{Plasticity rules:} ~~~~ \Delta \w = \eta z_t \x_t ~~.
\end{align}
The neural dynamics above describe a simple feedforward NN with inputs $\x_t$ and $y_t$. 
The input $\x_t$ is weighted by the synaptic connections $\w$, while the input $y_t$ has a fixed neural connection to the output neuron $z_t$. 
The update rule is a simple Hebbian one for $\w$ with pre-synaptic activities, $\x_t$, and post-synaptic activities $z_t$ as defined by Eq.~\eqref{eq:basic_hebbian_rule} in Sec.~\ref{sec:background}. 
Introducing the dual variable $z_t$ leads to neural dynamics and Hebbian rules. The biological realism of such an ``error" computing neuron is mentioned in Sec.~\ref{sec:Discussion}.

\subsection{Support Vector Machine}\label{subsec:SVM}

In the following, we consider the SVM problem introduced by \citet{cortes1995support} and formulated as follows
\begin{align}
    \min_{\w\in \R^{n}} \frac{1}{2} \|\w\|^2 ~~\text{s.t.} ~~ y_t \w^\top \x_t \geq 1 ~.
\end{align}
%

\paragraph{Dual objective function.}
The dual objective function for SVM is the following
\begin{align*}
&\max_{z_t \geq 0} - \frac{1}{2}\Tr (\chi^\top \chi \Z^\top \Z) + \left (\sum_t z_t ~~-\frac{\kappa}{2} z_t^2 \right)~~,
\end{align*}
with $\chi_t = y_t \x_t$. 
An important difference with the objective Eq.~\eqref{eq:dual_LR}, is that SVM considers the product of kernels, $\X^\top \X  \Y^\top \Y$ rather than their difference as in Eq.~\eqref{eq:MM_RR}. 
Nonetheless, the dual variables still account for some form of error.

\paragraph{Architecture, plasticity, and neural dynamics.}

The neural dynamics differ as it is a ``function" of the loss, not the regularizer 
\begin{align}
 &\text{Neural dynamics:} ~~ z_t =\frac{1}{\kappa} [1 -  y_t \w_t^\top  \x]_+ 
\\ & \text{Hebbian update:} ~~ \Delta \w =\eta z_t y_t \x_t  
\end{align}
This dual formulation of the update gives a simple version of the passive-aggressive algorithm introduced by \citet{crammer2006online}. 

The neural architecture arising from the primal-dual problem is that of a feedforward NN with rectified units. 
The interesting aspect of the architecture is that the output is a function of the product $y_t \x_t$, unlike in the regression problem Sec.~\ref{subsec:RR}. 
This is also the case for the Hebbian update rule for the weights $\w$, with pre-synaptic input $\x_t$, post-synaptic output $z_t$, and $y_t$ possibly operating as a modulating factor. 
Such an interaction can be speculated in the cerebellum, where this learning mechanism could be attributed to the interaction between Purkinje cells, $z_t$, parallel fibers $\x_t$, and climbing fibers $y_t$ \cite{moldwin2020perceptron}. 
SVM is also a candidate for odor classification in the mushroom body \cite{muezzinoglu2008artificial,nowotny2012equivalence}.

\paragraph{Connection with Similarity Matching.}
Again, as in Ridge regression, the synaptic weights, $\w$, arise from the kernel-to-kernel relationship between samples and neural output variable, $\Tr(\chi^\top\chi \Z^\top \Z)$. 
Maximization replaces minimization, and the input kernel considered here is the kernel product between labels and samples.

\subsection{Logistic Regression}

Another prevalent classification model is that of Logistic Regression (LR) \cite{cox1958regression}. The objective function for LR is that of binary cross-Entropy or Logloss. 
The LR model is 
\begin{align}
p(y = \pm 1 | \x, \w) = \frac{1}{ 1 + \exp(-y\w^\top \x)}~~.
\end{align}

\paragraph{Dual objective function.} 
In this context, \citet{jaakkola1999probabilistic} introduced the dual algorithm for LR as 
\begin{align}\label{eq:dual_LR}
R(\z)&= - \frac{1}{2}\sum_{t, t'} z_{t} z_{t'} y_{t} y_{t'} \x_{t}^\top \x_{t'} + \sum_t F(z_t ) ~, \nonumber
\\ &= - \frac{1}{2}\Tr(\chi^\top \chi \Z^\top\Z) + \sum_t F(z_t ) ~,
\end{align}
with $F(z) = - z \log(z) - (1-z)\log(1-z)$, and $\chi$ defined as in Sec.~\ref{subsec:SVM}.

\paragraph{Architecture, plasticity, and neural dynamics.}
Given the formulation of the dual objective function for LR, Eq.~\eqref{eq:dual_LR}, the update rules and neural dynamics are the following
\begin{align}
&\text{Neural dynamics:} ~~ \tfrac{d z_t(\gamma)}{d\gamma} = y_t \w_t^\top \x_t - F'(z_t(\gamma)) ~,
\\ &\text{Hebbian update:} ~~ \Delta \w =\eta z_t y_t \x_t  ~,
\end{align}
with $z_t\in [0,1]$. 
Here the neural activity is obtained as a fixed point of the neural dynamics. 
Such neural dynamics in the context of unsupervised learning can also be found in \cite{rozell2008sparse,obeid2019structured}.

The neural dynamics, as a function of the loss, differs from that of the SVM above. 
Here the neural activity is bounded between 0 and 1, as is expected from a sigmoid function. 
However, the update rule is the same. 
This is an example where more than knowing the update rule is required for distinguishing the underlying task being performed. 
However, with the power of the duality framework, looking at the neural dynamics in concert with the update rule should help in reverse-engineering neural computation.

\subsection{Multiplicative rules from Entropy regularization}

We have emphasized the importance of the $L_2$ regularization. However, other types of regularization are also useful. 
In the maximum entropy framework for density estimation, one seeks a density vector $\w$ that maximizes the relative entropy as 
\begin{align}
\hat{\w} = \argmax{\w \in \mathbb{R}^n_+ } g(\w) ~ \text{with} ~ y_t = \w^\top \x_t,~ \forall t \in \{1, n\}.
\end{align}
The regularization condition, $g(\cdot)$ is formulated as 
\begin{align}
g(\w) = \sum_{k=1}^n w_k \ln \left( \frac{w_k}{\mu_k} \right) ~~ , ~~ w_k\geq 0 ~,
\end{align}
which is the unnormalized entropy regularization. 
It has been popular in the online learning literature for leading to exponentiated gradient rules \cite{kivinen1995additive}. 
This learning problem has for dual variables neural dynamics, and update rules the following 
\begin{align}
&\text{Neural dynamics:} ~~\tfrac{dz_t(\gamma)}{d\gamma} = (y_t - \w^\top \x_t) - z_t(\gamma), 
\\ &\text{Hebbian update:} ~~~~ \hat{w}_{k,t+1} \propto \hat{w}_{k,t} \exp \left( z_{t} \x_{t}\right)  ~.
\end{align}
We can view this update as a multiplicative Hebbian rule. 
As the task is closely related to a regression model, the neural architecture and neural dynamics are unsurprisingly the same as that of RR. 
However, the difference lies in the update rule, determined by the regularization term. 
Indeed, the update is multiplicative. 
%
%
Again, the duality framework with constraints on the neural dynamics and the update allows us to narrow down the computation performed in specific neural components.

\section{Connection between the duality principle and biologically learning algorithms}\label{sec:Connection}

Based on the different examples given in Section~\ref{sec:NNandPlasticity}, we highlight how looking through the framework of duality can enlighten how we see neural plasticity and possibly reverse-engineer some observed anatomical and physiological observations.

\subsection{Sparse or dense dual variables} 

The dual variables in the supervised setting had the property of accounting for the error of the current model when evaluated on a new sample. 
This starkly contrasts with the type of dual variables mentioned in the SM framework, which is more unsupervised learning focused. 
Indeed, in an unsupervised setting, the dual variables account instead for the alignment between the sample and the current estimate. 

This has important implications for the nature of the update that can be tested in practice. 
In practice, supervised learning models will show sparse updates in time, while similarity-based models will show dense updates in time. 
Indeed, in unsupervised models, a better update typically results in a larger update. In contrast, a correct prediction does not require an update in supervised learning. 
Specifically, the dual variables $z_t$ become inactive in supervised learning once the optimal hyperplane is found for linearly separable problems.

\subsection{Additive or multiplicative updates}

From an ML perspective, we mentioned additive and multiplicative updates in the previous section. 
The distinction also exists in neuroscience as they correspond to two types of synaptic plasticity in the brain. 
Additive plasticity refers to changes in the synaptic strength independent of the synapse's current strength. 
In contrast, multiplicative plasticity refers to changes proportional to the synapse's current strength.

In the brain, additive plasticity is considered important for regulating the overall level of synaptic activity in a NN. 
On the other hand, multiplicative plasticity is important for regulating the relative strength of different synapses in a network \cite{loewenstein2011multiplicative}. 
This helps maintain the synapses' relative strength and prevent the weaker synapse from becoming saturated.

We observed that such rules could be employed for the same regression task but imply different regularization. 
This dichotomy is what distinguishes, for example, perceptron-like models from boosting-type models \cite{kivinen1999boosting,zemel2000gradient}. 
Entropy regularization is also a type of regularization often present in information-theoretic formulations. 
For example, in the recent work of \citet{halvagal2022combination}, they observe ``multiplicative" update rules such as the BCM rule. 
Similarly, for the work of \cite{bahroun2022similaritypreserving}.

Ultimately, both additive and multiplicative plasticity is essential, and in this work, we highlighted that a choice of learning rule can be directly traced back to the choice of the regularizer.  
We showed that the Representer theorem and duality allow reverse-engineering of the objective function and understanding of the neural circuit's task.

\subsection{$L_1$ penalty on the weights $\w$}

%
An important type of regularization that we have not covered yet that of sparsity-inducing regularization. 
From an ML perspective, the sparsity constraint is integral to the LASSO method. 
LASSO (least absolute shrinkage and selection operator) \cite{tibshirani1996regression,donoho1994ideal} is a regularized linear regression with $L_1$ penalty stated as 
\begin{align}\label{eq:LASSO}
\min_{\w\in\mathbb{R}^n} \sum_{t=1}^T ( y_{t} - \w^\top \x_t )^2  + \lambda \sum_{i=1}^n | w_i| ~~.
\end{align}
The LASSO method is often used in practice for feature selection, to select a few valuable and ignore the many that are not. 
We refer the reader to the work of \citet{bach2011convex} for an overview of convex optimization with sparsity-inducing norms. 
%


\paragraph{Sparse connections in biological NNs.}
Synaptic weights in the brain are sparse. Indeed, only a small subset of the connections between neurons in the brain exists \cite{song2005highly}, meaning that the networks are sparsely connected.

This has to be distinguished from the more common sparsity at the level of neural activity, which has been broadly studied as part of the theory of sparse coding \cite{olshausen1997sparse}, and sparse dictionary learning \cite{aharon2006k}. 
Biologically plausible implementations in the SM framework have also been studied \cite{pehlevan2014hebbian,bahroun2017online,lipshutz2022single}.

Connection sparsity is important in neuroscience for at least two reasons. 
Firstly, a sparsely connected network is considered important because it allows the brain to perform complex computations with relatively few active connections, reducing energy consumption and processing time. 
Secondly, it can be thought that the brain might be performing a form of a sparse mixture of experts, where not all features, i.e., upstream information, are relevant, allowing the extraction of only relevant information. 
In other words, synaptic weight sparsity is a form of neural efficiency. 
%

\paragraph{Plausible implementation.}
The work of \citet{connor2015biological} showed that by using a different prior than the Laplace prior, which they describe as ``offsetting or raising" the Laplacian prior distribution, they could propose a biological algorithm for Bayesian LASSO.  
As a result, they observed that the learning rule is now affected by a different learning rate, which in our framework can be represented by a set of dual variables. 
%
%
Interestingly, they argued that raising the Laplacian prior led to updated rules that better match the limited loss of association seen between days in the human data of the perceptual expertise task. 
%

\section{Discussion of existing methods}\label{sec:Discussion}

In this section, we discuss recent developments that rely on duality and some of the outlined elements in Sec.~\ref{sec:NNandPlasticity}. 
These works develop online algorithms for deep learning and biologically plausible alternatives to BP.

\subsection{Local learning rules in Deep Learning}\label{eq:local_DL}

Various recent contributions have been developed based on duality in deep learning. 
In particular, one of the earlier works is the Method of Auxiliary Coordinates (MAC) from \citet{carreira2014distributed}

The idea is the following, training a fully-connected NN with $L$ hidden layers consists of minimizing the loss $\mathcal{L}(\y, \sigma(\W, \x_L))$ involving a nested function 
\begin{align}
f(\W, \x_L) = f_L(\W_L, \sigma_{L-1}( \W_{L-1}, \ldots, \sigma_1(\W_1, \x_L) ) ~,
\end{align}
which can be re-written as constrained optimization: 
\begin{eqnarray}
\min_{\W} \sum_{t=1}^n  \ell (\y_t, \a_t^L, \W_{L+1})~~ , ~~\text{where}~~ \a_t^l = \sigma_l (\c_t^l),~~ \\ 
\text{s.t.} ~~ \c_l^t = \W_l \a^{l-1}_{t}, ~\forall l \in \{1, L\} ~~\text{and}~~ \a_t^0 = \x_t. \label{eq:nested_constraints}
\end{eqnarray}
The NN becomes a succession of regularized objectives that we detailed above, which are types of RR. 
The generality of the duality approach can be seen in the fact that a lot of novelties introduced by \citet{carreira2014distributed} could be carried away to the work of \citet{choromanska2019beyond} by changing the objective function. 
And the work of \citet{choromanska2019beyond} recovering Hebbian-like learning rules and local quantities is a direct consequence of the Representer theorem and choice of the regularizer.

Other objective functions, which directly consider the notion of similarities for optimizing NNs as in \cite{pogodin2020kernelized,ma2020hsic}, have been proposed and, as part of the framework, lead to local learning rules or alternatives to BP. 
Although the works mentioned above show some biological plausibility, they are arguably separated from the more neuroscience-driven approaches, which we detail in the following section.

\subsection{Local learning for neuroscience}

In the last few years, various attempts have been made at building biologically plausible alternatives to BP. 
Interestingly, the approaches were similar to those in Sec.~\ref{eq:local_DL}. 
Although not always explicitly mentioned, those models fall in the category of nested objectives, where each layer can be phrased as in Sec.~\ref{sec:NNandPlasticity}.

Those approaches focused on building or mapping the resulting component onto anatomical and physiological observations. 
The two approaches detailed below are to either consider the dual variables as quantities computed in a dendritic compartment of pyramidal-like cells or to have them computed by error neurons.

\paragraph{Dendritic computation of error signal.} 

In their work, \citet{sacramento2018dendritic} proposed representing feedback error signals in dendritic compartments. 
Similar ideas were also explored in the work of 
\citet{golkar2020simple} and \citet{meulemans2021credit}, each adding various valuable biological components to their models. 
It is important to note that all these models have in common to impose regularization properties to their models so that local (dual) variables exist and can be interpreted as biophysical quantities.

This line of work has also been furthered in \cite{meulemans2022least,zucchet2022beyond} by introducing bilevel optimization and dual variables to produce quantities that can be computed locally. 
We also refer the reader to work on deep equilibrium models \cite{bai2019deep} for completeness.  

\paragraph{Error and inter-neurons.} 
Another line of work focused on the predictive coding theory that local error neurons can use two distinct populations of neurons \citet{whittington2017approximation}. 
Error neurons measure deviations of the actual state of prediction neurons from the expected state. This is very similar to what the dual formulation offers. 
Indeed, the probabilistic formulation of predictive coding makes it similar to regularized generalized linear models and that of \citet{carreira2014distributed}.

More recently, the work of \citet{golkar2022constrained} has shown that the introduction of optimization constraints related to whitening allowed the introduction of interneurons alleviating various limitations on the biological implementations of the predictive framework. 
The reader can find other discussions on the possible implementation of the PC framework in the work of \citet{ororbia2019biologically}.

\section{Conclusion}


In this work, we aimed to show that local learning rules, the Hebbian principle, and the Representer theorem, although coming from different fields, can be considered similar concepts and should be used in concert. 
We have provided well-known examples of ML algorithms that can be naturally phrased as biologically plausible. We also emphasized how the variety of approaches covered by Representer can account for biologically relevant features.

In conclusion, adopting a viewpoint based on dual variables and the Representer theorem during the development of algorithms offers a twofold advantage. 
Firstly, it enables the creation of biologically plausible algorithms that align with observable physiological and anatomical observations. 
Secondly, it facilitates the reverse-engineering process, allowing for the interpretation and mapping of local variables to canonical components found in the brain.
Researchers are bridging the gap between computational models and biological systems by leveraging dual variables, paving the way for advancements in our understanding of neural processes.

\bibliography{example_paper.bib}
\bibliographystyle{icml2023}

\end{document}